\documentclass[letterpaper, 10 pt, journal, twoside]{IEEEtran}

\usepackage{amsmath,amsfonts}
\usepackage{algorithmicx}
\usepackage{algorithm}
\usepackage{array}
\usepackage[caption=false,font=normalsize,labelfont=sf,textfont=sf]{subfig}
\usepackage{textcomp}
\usepackage{stfloats}
\usepackage{url}
\usepackage{verbatim}
\usepackage{graphicx}
\usepackage{cite}
\usepackage{graphics}
\usepackage{bm}
\usepackage{multirow}
\usepackage[algo2e]{algorithm2e}
\usepackage{makecell}
\usepackage{amssymb}
\usepackage{booktabs}
\usepackage[margin=2cm]{geometry}
\usepackage{xcolor}
\usepackage[compatible]{algpseudocode} 
\usepackage{threeparttable}

\usepackage{amsthm}

\usepackage{xurl}

\newcommand{\scenario}[1]{{\smaller \sf#1}\xspace}
\newcommand{\flores}{\scenario{FLORES}}

\IEEEoverridecommandlockouts                              

\begin{document}

\title{
A Reconfigured Wheel-Legged Robot\\ for Enhanced Steering and Adaptability
}


\author{Zhicheng Song$^*$, Jinglan Xu$^*$, Chunxin Zheng, Yulin Li, Zhihai Bi, and Jun Ma, \textit{Senior Member, IEEE}%
\thanks{$^{*}$indicates equal contribution.}%
\thanks{Zhicheng Song, Jinglan Xu, Chunxin Zheng, Yulin Li, Zhihai Bi, and Jun Ma are with the Robotics and Autonomous Systems Thrust, The Hong Kong University of Science and Technology (Guangzhou), Guangzhou 511453, China (e-mail: zsong469@connect.hkust-gz.edu.cn; jxu367@connect.hkust-gz.edu.cn; czheng739@connect.hkust-gz.edu.cn; yline@connect.ust.hk; zbi217@connect.hkust-gz.edu.cn; jun.ma@ust.hk).}%
}

\maketitle


\begin{abstract}
Wheel-legged robots integrate leg agility on rough terrain with wheel efficiency on flat ground.
However, most existing designs do not fully capitalize on the benefits of both legged and wheeled structures, which limits overall system flexibility and efficiency.
We present \flores, a novel wheel-legged robot design featuring a distinctive front-leg configuration that sets it beyond standard design approaches.
Specifically, \flores replaces the conventional hip-roll degree of freedom (DoF) of the front leg with hip-yaw DoFs, and this allows for efficient movement on flat surfaces while ensuring adaptability when navigating complex terrains.
This innovative design facilitates seamless transitions between different locomotion modes (i.e., legged locomotion and wheeled locomotion) and optimizes the performance across varied environments.
To fully exploit \flores's mechanical capabilities, we develop a tailored reinforcement learning (RL) controller that adapts the Hybrid Internal Model (HIM) with a customized reward structure optimized for our unique mechanical configuration. 
This framework enables the generation of adaptive, multi-modal locomotion strategies that facilitate smooth transitions between wheeled and legged movements. 
Furthermore, our distinctive joint design enables the robot to exhibit novel and highly efficient locomotion gaits that capitalize on the synergistic advantages of both locomotion modes.
Through comprehensive experiments, we demonstrate \flores's enhanced steering capabilities, improved navigation efficiency, and versatile locomotion across various terrains.
The open-source project can be found at \url{https://github.com/ZhichengSong6/FLORES}.

\end{abstract}

\section{INTRODUCTION}

\IEEEPARstart{W}{heel-legged} robots have emerged as a promising solution for diverse locomotion tasks, which seamlessly integrate the agile walking capabilities of legged robots for navigating rough terrain with the high-efficiency motion of wheeled robots on flat surfaces~\cite{hutter2017anymal,kim2019highly,KeepRolling, ETHANYmalonWheel}. These hybrid systems have shown remarkable potential in applications such as exploration, rescue, and delivery~\cite{10.1002/rob.21677,park2023wave}. 
However, despite these significant advancements, two fundamental challenges persist in fully capitalizing on the advantages of both legged and wheeled locomotion: wheel-leg mechanical design and control strategies.  

\begin{figure}[tp]
    \centering
    \includegraphics[width = 0.47\textwidth]{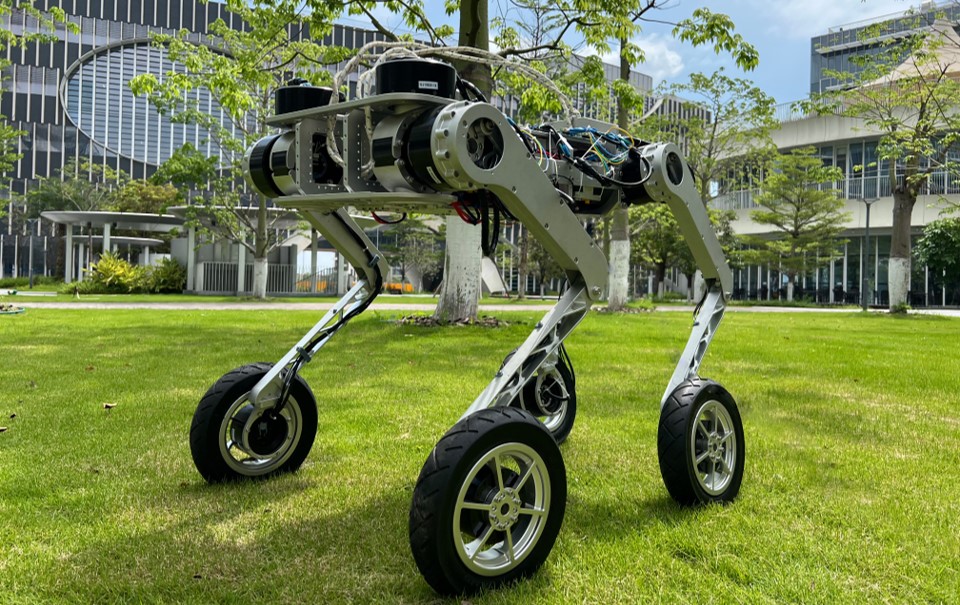} 
   \caption{\textbf{\flores}: A reconfigured wheel-legged robot designed for enhanced steering and adaptability across diverse terrains.}
    \label{fig:robot_stand_in_the_wild}
    \vspace{-0.4cm}
\end{figure}

The structural design of current wheel-legged robots generally falls into two categories, each with its own limitations. 
The first category treats the legged structure as an active suspension system.
This design often results in fewer actuators~\cite{ni2022parameters} or limited joint ranges~\cite{klemm2019ascento}, restricting the robot's performance in environments that require rapid or dynamic mode switching.
Representative systems such as SherpaTT further demonstrate actively articulated suspension designs in Mars-analog field tests~\cite{cordes2018design}.
Some designs within this category, such as~\cite{reid2016actively}, utilize actuators that provide low output speeds, while others incorporate harmonic reducers~\cite{kashiri2019centauro}, which offer high torque at the expense of speed. 
Combined with a heavy leg structure, this makes the system unsuitable for high-dynamic hybrid locomotion.

The second category involves directly modifying existing legged platforms by incorporating wheels~\cite{KeepRolling, ETHANYmalonWheel}. 
In this configuration, the wheel serves as a supplementary feature rather than a core design element. While it enhances straight-line locomotion, it is less adaptable for more intricate movements, such as turning or traversing cluttered environments~\cite{lee2024learning, cui2021learning, zhao2021design}, as most complex maneuvers beyond straight-line motion require leg lifting and repositioning, during which the increased wheel weight consistently imposes a negative impact on the robot's agility and energy efficiency, particularly when navigating unstructured or rough terrains.

On the control side, while model-based control strategies have achieved notable success in legged robots~\cite{8276298, 8460731}, they do not naturally generalize to wheel-legged robots due to their heavy reliance on predefined contact sequences~\cite{8260889, 8793865} that mimic biological gaits~\cite{8593885}. 
However, hybrid wheel-legged systems lack direct biological analogs, leading these traditional model-based methods to treat wheeled and legged locomotion in isolation~\cite{KeepRolling, ETHANYmalonWheel} and sacrificing system flexibility. 
In contrast, reinforcement learning (RL) has emerged as a dominant paradigm for generating wheel-legged motion, demonstrating superior adaptability in learning hybrid locomotion patterns without predefined sequences~\cite{lee2024learning, zhu2024whleaper, chamorro2024reinforcement}.

We propose \flores, an innovative wheel-legged robot that differs from conventional designs. 
As shown in Fig.~\ref{fig:hardware_BOM}, we reconfigure the front leg by incorporating a hip-yaw joint, which significantly enhances the robot's steering capabilities. 
The hip-yaw joint acts as a comprehensive steering module, enabling efficient locomotion on relatively flat surfaces without frequent leg lifting, similar to a vehicle steering mechanism. 
This design improves both steering capabilities and adaptability in complex environments while maintaining the linear-motion efficiency. 
Moreover, the unmodified rear legs preserve the robot's ability to navigate and stabilize on more complex, unstructured terrains, ensuring versatility across various environments. 
To optimize the capabilities of this hybrid configuration, we modify the Hybrid Internal Model (HIM) framework \cite{long2023him} to align with our distinct mechanical design.
We customize the reward structure tailoring to \flores's innovative features, which places particular emphasis on steering, dynamic adaptability, and the ability to navigate complex terrains.
As demonstrated in our experiment, our method enables dynamic optimization of locomotion across a wide range of challenging scenarios, showcasing the synergy between \flores's innovative design and its control framework.

To summarize, the main contributions of this paper are as follows:
\begin{itemize} 
\item {New Wheel-Legged Robot Platform: We introduce \flores, a wheel-legged robot platform designed with an innovative front-leg configuration that synthesizes the speed and efficiency of wheels in conjunction with the adaptability and maneuverability of legs, enabling versatile locomotion across various terrains.}

\item {Customized RL Control Framework for \flores: We develop a customized RL policy tailored to \flores that leverages the synergistic advantages of wheeled and legged locomotion while ensuring smooth transitions between these modes.}

\item {Demonstration of Superior Performance: We demonstrate \flores's superior steering abilities, increased navigation efficiency, and adaptability across diverse terrains, particularly in disturbance-rich and cluttered scenarios requiring frequent reorientation.}

\item {Open-Source Hardware and Software Implementation: We provide comprehensive open-source mechanical and electrical designs, along with the complete controller implementation.}
\end{itemize}

\section{RELATED WORK}
\subsection{Wheel-Legged Robot Platforms}
Existing wheel-legged robot platforms can generally be categorized into two design approaches regarding the integration of wheels:

\noindent \textbf{Wheels with Active Suspension.} Bipedal designs such as Ascento~\cite{klemm2019ascento} provide efficient movement on flat terrain, but due to their limited number of actuators and restricted joint range, they can hardly walk and are primarily designed for jumping to overcome obstacles. 
On the other hand, quadrupedal designs like Momaro~\cite{klamt2017anytime} and Centauro~\cite{kashiri2019centauro} can simultaneously drive and walk. 
Nevertheless, Centauro's performance is constrained by harmonic reducers in its actuators and heavy leg structures, which limit its capacity for high-dynamic locomotion.
While harmonic reducers provide high torque, they suffer from low output speeds and poor resistance to continuous impacts, reducing the robot's agility and overall performance in dynamic environments.
Although series elastic actuation is employed to enhance structural integrity, it introduces significant nonlinearity in motor control. 
Similarly, Momaro utilizes servo motors as actuators, which also deliver low output speeds.

\noindent \textbf{Wheels as Add-on Features.} The DRC-HUBO+~\cite{lim2017robot} is a bipedal robot that incorporates an additional wheel structure.
However, it struggles with hybrid locomotion, as reconfiguration is necessary when switching between walking and driving modes.
A more advanced design is ANYmal~\cite{KeepRolling, ETHANYmalonWheel}, which operates in walking, driving, or hybrid modes. 
Built on the legged version of ANYmal, a state-of-the-art quadrupedal robot platform, this design offers improved versatility.
The wheel structure excels at straight-ahead travel by exploiting pure rolling, which is energy-efficient. However, in complex environments, essential movements like turning and lateral movement that require frequent foot lifting could result in reduced energy efficiency.

\subsection{Control Strategies for Wheel-Legged Robot}
Control strategies for wheel-legged robots broadly fall into two categories: model-based and learning-based control.

\noindent
\textbf{Model-based Control.} 
ANYmal uses model-based approaches like Model Predictive Control (MPC) \cite{ANYmalWheel_WBCMPC} and Whole-Body Control (WBC) \cite{ETHANYmalonWheel, KeepRolling} to optimize joint velocities and ground reaction forces based on dynamic models and constraints. 
However, these methods require accurate robot models and predefined contact sequences, limiting their adaptability in dynamic or rough terrain.
The LQR-assisted WBC \cite{klemm2020lqr}, applied to simpler structures like Ascento, eliminates the need for predefined contact sequences but it sacrifices applicability in complex hybrid locomotion scenarios.
Despite their successes, model-based approaches face challenges in obtaining accurate robot models and struggle with adaptability in dynamic environments, even when incorporating online gait generators \cite{ANYmalWheel_WBCMPC}.

\noindent
\textbf{Learning-based Control.} Recent advances in massively parallel RL simulations \cite{makoviychuk2021isaac} have significantly reduced the time required for policy development, making RL increasingly appealing for tasks that involve diverse terrains or unpredictable obstacles, where traditional model-based approaches may struggle to maintain optimal performance.
For instance, ANYmal employs a recurrent neural network-based policy to facilitate robust and versatile locomotion~\cite{lee2024learning}. 
Additionally, Whleaper, a bipedal wheel-legged robot, applies the Proximal Policy Optimization (PPO) algorithm to facilitate reliable walking and jumping locomotion~\cite{zhu2024whleaper}. 
Similarly, PPO with an asymmetric actor-critic architecture enables blind stair climbing for various wheel-legged robots~\cite{chamorro2024reinforcement}. 
Recent works also explore learning-enhanced disturbance compensation in structured controllers. For example, NeuroMHE~\cite{NeuroMHE} combines recurrent networks with moving horizon estimation to infer latent disturbances, while L1-MPC~\cite{L1-MPC} integrates L1 adaptive control into MPC formulations for improved robustness against unmodeled dynamics.

In our work, we adopt the HIM framework \cite{long2023him}, which integrates classical Internal Model Control (IMC) principles with the PPO algorithm. 
This framework estimates environmental disturbances, such as ground friction and terrain elevation, using only the robot's proprioceptive data from inertial measurement units (IMUs) and joint encoders, thereby eliminating the need for additional sensors.

\section{METHOD}

\subsection{Hardware Design} 
\flores is designed for versatile locomotion across terrains, exploiting both legged and wheeled structures\footnote{All the hardware design and a datasheet for all electronics can be found in our project website.}.



\begin{figure}[t]
    \centering
    \includegraphics[width = 0.5\textwidth]{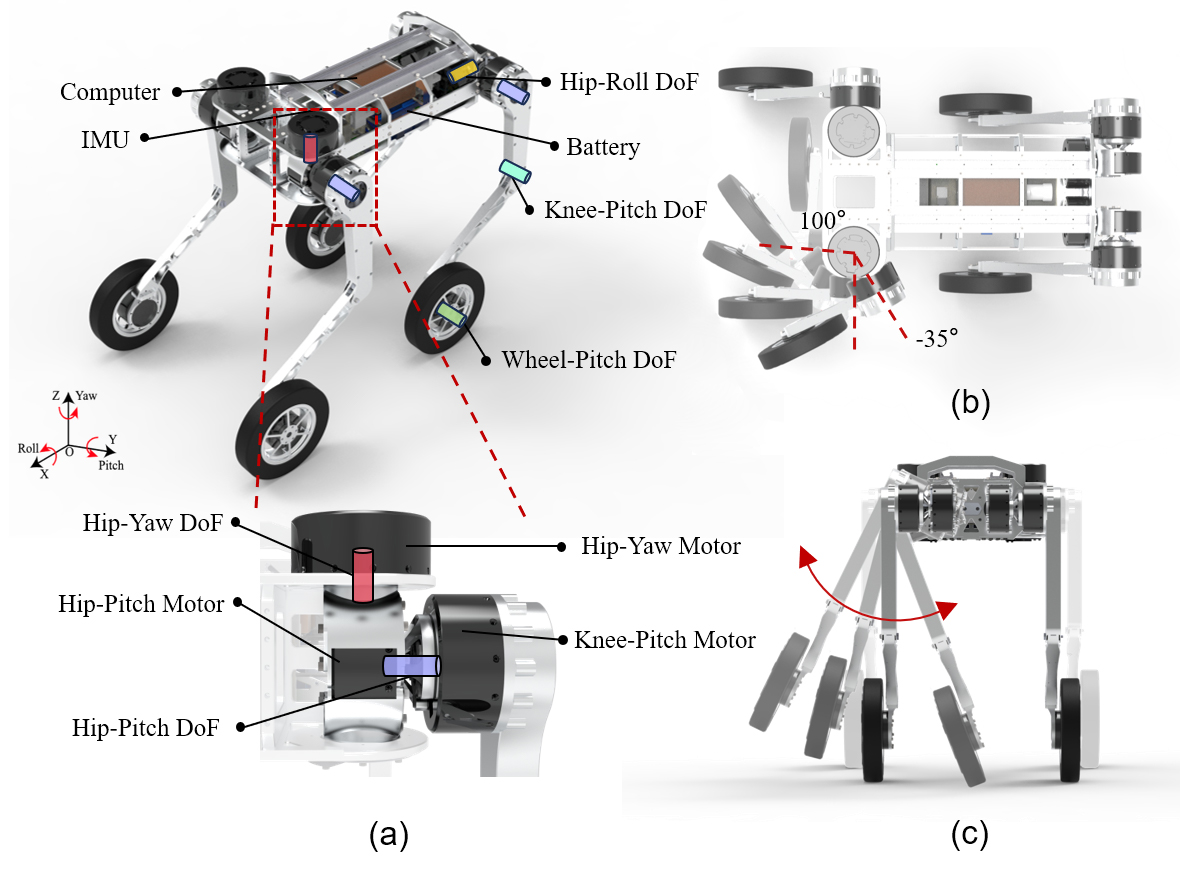}
    \caption{Overview of mechanical design: (a) Illustration of the joint DoF design for \flores, along with the positioning of electrical components. The magnified view on the lower side shows the joint layout of the front left leg. (b) Range of motion for the front hip joint. (c) Range of motion for the rear hip joint.}
    \label{fig:hardware_BOM}
    \vspace{-12pt}
\end{figure}

\noindent
\textbf{Mechanical Design.} \flores has a total of 16 degrees of freedom (DoFs) with 4 DoFs per leg, as shown in Fig. \ref{fig:hardware_BOM}(a).
Each leg is equipped with 3 motors near the torso: two motors provide hip DoFs (hip-yaw \& pitch for front legs/ hip-roll \& pitch for rear legs), and one for actuating the knee-pitch DoF. 
Additionally, a distal motor drives each wheel.
The hip-pitch and knee-pitch motors are co-axially located and the knee-pitch DoF is actuated through a parallel four-bar linkage mechanism, which reduces leg inertia, and all other DoFs are directly actuated by motors.
Such parallel linkage mechanisms are commonly adopted in series-parallel hybrid robotic systems for improved structural and dynamic performance~\cite{KUMAR2020102367}.

In particular, as demonstrated in Fig. \ref{fig:hardware_BOM}(b), the front legs feature a hip-yaw DoF, which acts as a steering module and significantly improves the robot's maneuverability on unstructured terrain. 
To obtain enough steering space for front legs, the distance between front hip joints is larger than the back hip joints, which can prevent the leg collide with the body when a big heading angle is needed.
The hip-yaw DoF has a range of motion from -35° to 100°, providing greater flexibility in adjusting the robot’s heading. 
This range improves agility, allowing the robot to quickly change directions—especially beneficial in tight or complex environments. 
Meanwhile, as shown in Fig. \ref{fig:hardware_BOM}(c), the rear legs with their typical hip-roll configuration maintain stability on unstructured terrain, ensuring the robot’s adaptability in various environments. 
As such, each front leg is equipped with motors for hip-yaw, hip-pitch, knee-pitch, and wheel actuation, while the rear legs retain the typical hip-roll configuration.

When navigating unstructured terrain, the modified hip-yaw DoF significantly enhances the robot's efficiency by allowing the front wheels to maintain ground contact as long as possible, thus enabling a greater reliance on the wheeled structure for generating various movements. 
On the other hand, the typical hip-roll DoF in the rear leg, while not as adept at maximizing the use of the wheeled structure, provides essential lateral force that ensures stability on uneven surfaces. 
This lateral force is crucial for maintaining balance, as a lack of it could easily lead to tipping over when traversing diverse terrains.


\noindent
\textbf{Actuators.} All joints use Motorevo brushless DC motors with 10:1 planetary reducers. The hip–yaw/roll/pitch and knee–pitch motors provide 32 N·m continuous torque, and the wheel motors 8 N·m.
Each motor is equipped with a single encoder to provide position, velocity, and current measurement feedback. 
Motor control is achieved via the EtherCAT communication protocol.
The relatively low reduction ratios of planetary reducers are chosen for two primary reasons: first, they facilitate high-speed motion, which is essential for dynamic locomotion; second, these quasi-direct drive motors provide relatively accurate torque feedback through current measurements, eliminating the need for additional sensors. Furthermore, planetary reducers are capable of withstanding significant unexpected impacts, which frequently occur during dynamic movement.

\noindent
\textbf{Manufacturing.} The body plates and legs of the robot are constructed from off-the-shelf aluminum 6061, while certain electronic mounts are 3D-printed using Polylactic Acid (PLA). 

\noindent
\textbf{Sensors and PC.} An IMU (HiPNUC-HI13R4) is mounted on the base, providing precise state measurements of the robot at 500 Hz. 
The main processing unit, a Piesia-TL500Z3AW-R110 with an Intel i7-1165G7 processor, executes the control policy at 50 Hz.

\subsection{Software Design}

\subsubsection{Overview}
\begin{figure*}[t]
    \centering
    \includegraphics[width = 0.95\textwidth]{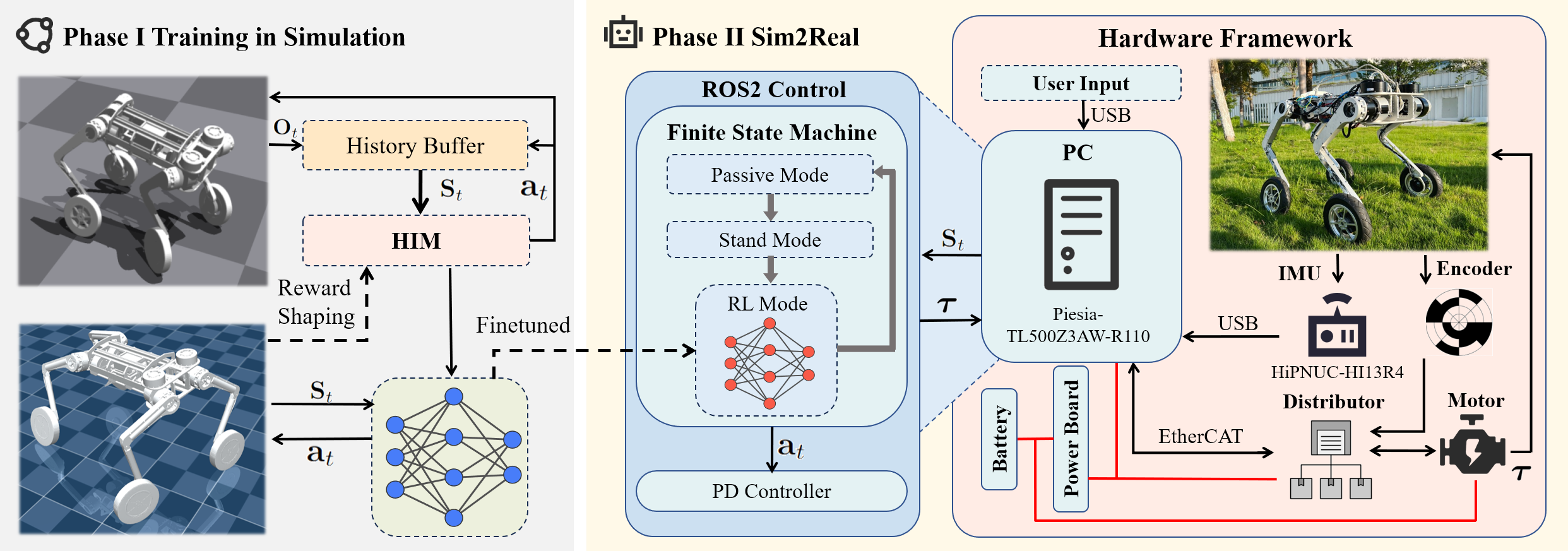}
    \caption{Overview of our system architecture and training pipeline. The training process consists of two phases: Phase I involves sim-to-sim transfer, where the HIM policy is first trained in Isaac Gym with reward shaping, and then is transferred and validated in MuJoCo as an intermediate step; Phase II performs Sim2Real deployment using ROS2 control framework to establish robot communications. The hardware framework (right panel) shows the onboard PC receiving user input and combining it with current observations to feed into the trained RL model, which outputs actions for robot control. Red lines indicate power connections, and black arrows show the information flow.}
    \label{fig:system_framework}
    \vspace{-12pt}
\end{figure*}

As illustrated in Fig.~\ref{fig:system_framework}, the software architecture is composed of two sequential phases to realize robust locomotion control on our wheel-legged robot. 
In Phase $\mathrm{I}$, training is conducted in Isaac Gym~\cite{makoviychuk2021isaac}, where the robot’s state is described by joint velocity, joint position, angular velocity, gravitational force, and the previous action. 
These inputs are processed by HIM, which integrates an encoder-based state representation with a customized reward function. 
The policy network is then optimized via PPO, providing target wheel velocities and joint positions as output commands.
In Phase $\mathrm{II}$, the learned policy is first fine-tuned in MuJoCo~\cite{todorov2012mujoco} under more realistic dynamics.
Subsequently, the final policy is deployed onto the real robot. 
Hardware feedback—including signals from the IMU and motor encoders are fed to the policy, which runs on an embedded compute unit.
By sending real-time torque commands, the robot achieves smooth sim-to-real transitions and robust performance in physical environments.

\subsubsection{Reinforcement Learning for \flores}

As shown in Fig.~\ref{fig:system_framework}, the task of the policy is to control the robot to follow the desired velocity $\mathbf{v}^{\text{cmd}} = [v_x^{\text{cmd}}, v_y^{\text{cmd}}, \omega_{z}^{\text{cmd}}]\in \mathbb{R}^{3}$, which indicates the longitudinal and lateral linear velocities, and angular velocity in the horizontal direction, respectively.

\noindent
\textbf{Observations.} 
In our RL control policy, the state vector \(\mathbf{S}_t \in \mathbb{R}^{689}\) contains 1 step current observation \(\mathbf{O}_t \in \mathbb{R}^{53}\) and 12 step history observation \(\mathbf{O}_{t-12:t-1} \in \mathbb{R}^{636}\). Each 53-dimensional observation vector comprises the following components:
\begin{itemize}
    \item The robot's angular velocity 
    \(\boldsymbol{ \omega }_{\mathrm{base}} \in \mathbb{R}^3\) in the body frame;
    \item The gravity vector \(\mathbf{g} \in \mathbb{R}^3\) in the body frame;
    \item A command vector  $\mathbf{v}^{\text{cmd}} = [v_x^{\text{cmd}}, v_y^{\text{cmd}}, \omega_{z}^{\text{cmd}}]\in \mathbb{R}^{3}$ describing the desired linear and angular velocities;
    \item A 12-dimensional vector of joint position errors, except for the wheels;
    \item A 16-dimensional joint velocity vector $\dot{\mathbf{q}} = [\dot{\mathbf{q}}_{\text {leg }}, \dot{\mathbf{q}}_{\text {wheel }}]$;
    \item The 16-dimensional action \(\mathbf{a}_{t-1} \in \mathbb{R}^{16}\) taken at the previous timestep.
\end{itemize}
These inputs constitute the observation \(\mathbf{S}_t\), which is then processed by the control policy to generate the appropriate control signals.

\noindent
\textbf{Actions.} 
The action \(\mathbf{a}_{t} \in \mathbb{R}^{16}\) contains two different categories of actions: \(\mathbf{a}^{\text{leg}}_{t} \in \mathbb{R}^{12}\) and \(\mathbf{a}^{\text{wheel}}_{t} \in \mathbb{R}^{4}\).
\(\mathbf{a}^{\text{leg}}_{t}\) is the bias between the desired joint angle and the default joint angle for 12 joints, except for the wheels. 
A fixed action scale \({k}\) is multiplied by the action to reduce output instability.
The desired joint angle is the sum of the default joint angle and the product of the scale factor and the action, which can be defined as 
\begin{equation}
    \mathbf{q}^{\text{des}}=\mathbf{q}^{\text{default}}+k\mathbf{a}^{\text{leg}}_t.
\end{equation}
The second part \(\mathbf{a}^{\text{wheel}}_{t} \in \mathbb{R}^{4}\) is the desired velocity vector for wheels since the velocity control is implemented on the wheel.

\noindent
\textbf{Reward.} 
Our reward design follows a standard Legged-Gym–style shaping~\cite{leggedGym}, with adjustments made solely to the weights for our specific platform.
Table \ref{tab:reward function} presents the reward functions for our robot. 
In this context, $e_v$ and $e_{\omega_z}$ represent the tracking errors for linear velocity and angular velocity, respectively, while the function $n(e)$ indicates whether the error $e$ is close to zero. 
The variables $v_z$, $\omega_{x}$, $\omega_{y}$, and $g^b_{xy}$ denote the base velocity along the $z$-axis, angular velocity around the $x$-axis, angular velocity around the y-axis, and the projected gravity in the $x$ and $y$ directions, respectively. 
Additionally, $h$, $h_d$ and $\mathcal{I}$ represent the height of the robot, the target height of the robot, and an indicator function that equals zero when the robot is in motion. 
The variable $\tau$ refers to the output torque of all the joints, while $\sigma_1$, $\sigma_2$, $\sigma_3$, $\sigma_4$, and $\sigma_5$ are user-defined coefficients for the various reward components.

\begin{table}[t]
\centering
\caption{Reward Function}
\begin{tabular}{rcr}
\toprule
Reward Terms&Definition&Weight\\
\midrule
Tracking Lin. Vel.  & $ \begin{cases} -|e_v|, & \text{if } n(e_v) \\ \exp\left(-\dfrac{|e_v|}{\sigma_{1}}\right), & \text{otherwise} \end{cases}$  &   8.0\\

Tracking Ang. Vel.  & $ \begin{cases}  -|e_{\omega_z}|, & \text{if } n(e_\omega) \\ \exp\left(-\dfrac{|e_\omega|}{\sigma_{2}}\right), & \text{otherwise} \end{cases}$  &   4.0\\

$z$-axis Lin. Vel.    &   $v_z^2$   &    -0.1\\

$x$-axis \& $y$-axis Ang. Vel.   &   $\omega_{x}^2+\omega_{y}^2$    &   -0.05\\

\midrule

Orientation &   $(\mathbf{g}_{xy}^b)^2$ &   -1.0\\

Base Height &   $\exp \left\{\frac{-|h-h_{d}|}{\sigma_{3}} \right\}$ &   2.0\\

Static Pose &   $\mathcal{I} \exp \left\{\frac{- |\mathbf{q}_{\text{leg}} - \mathbf{q}_{\text{leg}}^{\text{default}}|^2}{\sigma_{4}} \right\}$ &   5.0\\

Dynamic Pose    &   $\exp \left\{\frac{- |\mathbf{q}_{\text{leg}} - \mathbf{q}_{\text{leg}}^{\text{default}}|^2}{\sigma_{5}} \right\}$    &   1.0\\

\midrule
Joint Acc.  &   $\left\|\ddot{\mathbf{q}}\right\|_2^2$  &   -2.5e-7\\

Joint Power &   $|\boldsymbol{\tau}||\dot{\mathbf{q}}|^T$   &   -5e-5\\

Torques &   $|\boldsymbol{\tau}|^2$    &   -5e-5\\

Action Rate &   $(\mathbf{a}_{t-1}-\mathbf{a}_t)^2$ &   -0.01\\

Smoothness  &   $(\mathbf{a}_{t}-2\mathbf{a}_{t-1}+\mathbf{a}_{t-2})^2$ &   -0.01\\
\bottomrule
\end{tabular}

\label{tab:reward function}
\vspace{-10pt}
\end{table}

\noindent
\textbf{Training.} We train the 4096 parallel environment for 18 hours with Issac Gym on a single A100 GPU.

\subsubsection{Sim2Real Transfer for \flores}


Before deployment, we address performance degradation caused by mismatches in physics engines, dynamic models, and environment configurations. To mitigate this sim-to-real gap, we apply domain randomization in Isaac Gym~\cite{makoviychuk2021isaac} to promote robustness under diverse conditions. We then validate and fine-tune the policy in MuJoCo~\cite{todorov2012mujoco}, preserving its performance in a new simulation environment while reducing risks in real-world experiments.

We further extend domain randomization to actuator and sensor uncertainties: we perturb motor strength and PD gains to accommodate hardware deviations, randomize external pushes (forces/velocities) to emulate interactions, and inject variable observation delays to reflect perception latency. The communication frequency between the motor driver and PC is fixed at 4 kHz to stabilize command timing and smooth actions.
Together, these strategies narrow the sim-to-real gap for our wheel-legged robot and maintain robust performance even in complex, unstructured environments.


\section{Experiment}\label{sec:experiments}

In this section, we present a comprehensive series of experiments to demonstrate the enhanced steering capabilities, improved navigation efficiency, and versatile locomotion across various terrains.
We first show \flores has the ability to navigate through various terrains with multi-modal locomotion with smooth transitions between wheeled and legged movement.
Additionally, we compute the Cost of Transport (CoT) for \flores and B2W, a state-of-the-art wheel-legged platform from Unitree Robotics, during various navigation tasks to quantify \flores's superior steering ability and navigation efficiency.
All comparisons use identical task definitions and command profiles.

\subsection{Performance on Various Terrains}

\begin{figure*}[tbp]
    \centering
    \includegraphics[width = 0.96\textwidth]{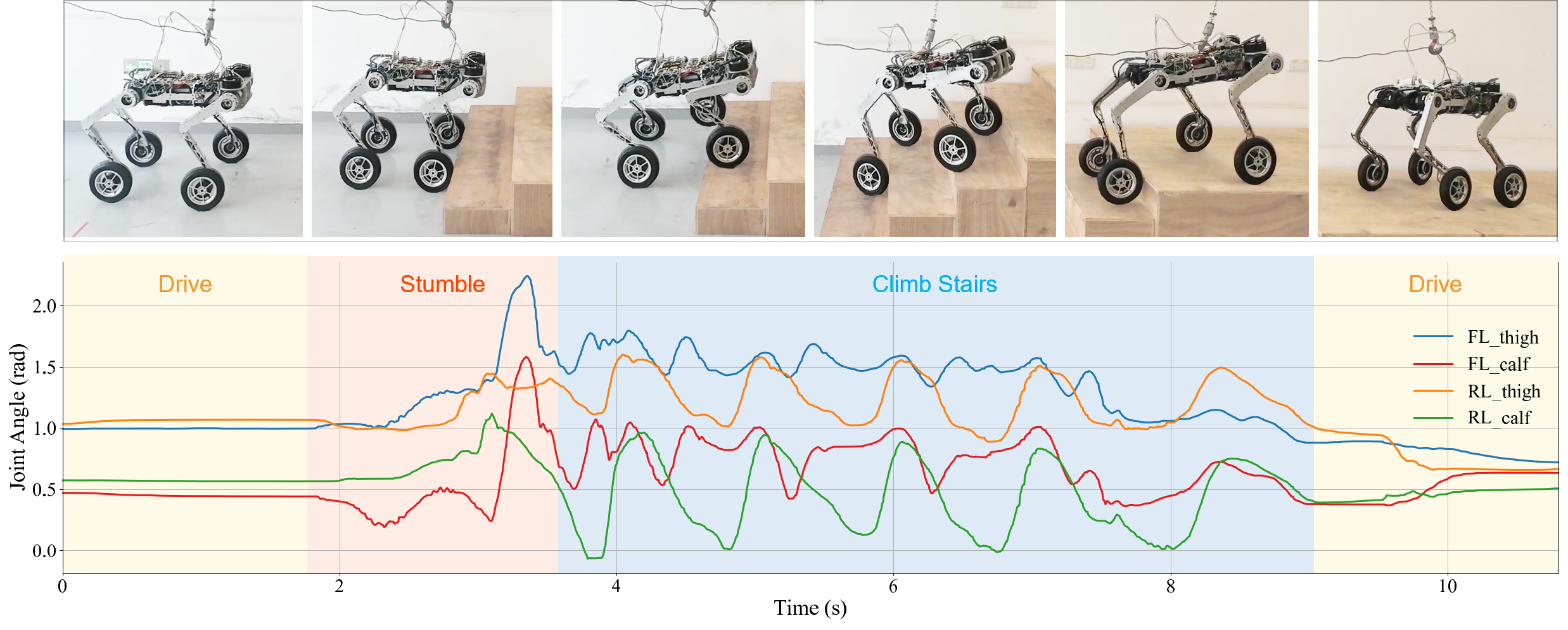}
    \vspace{-10pt}
    \caption{The top figure illustrates the progress of \flores as it climbs up the stairs, while the bottom figure depicts the joint angles of the thigh and calf joints for both the front-left and rear-left legs.}
    \label{fig:staris}
    \vspace{-8pt}
\end{figure*}

\begin{figure}[t]
    \centering
    \includegraphics[width = 0.48\textwidth]{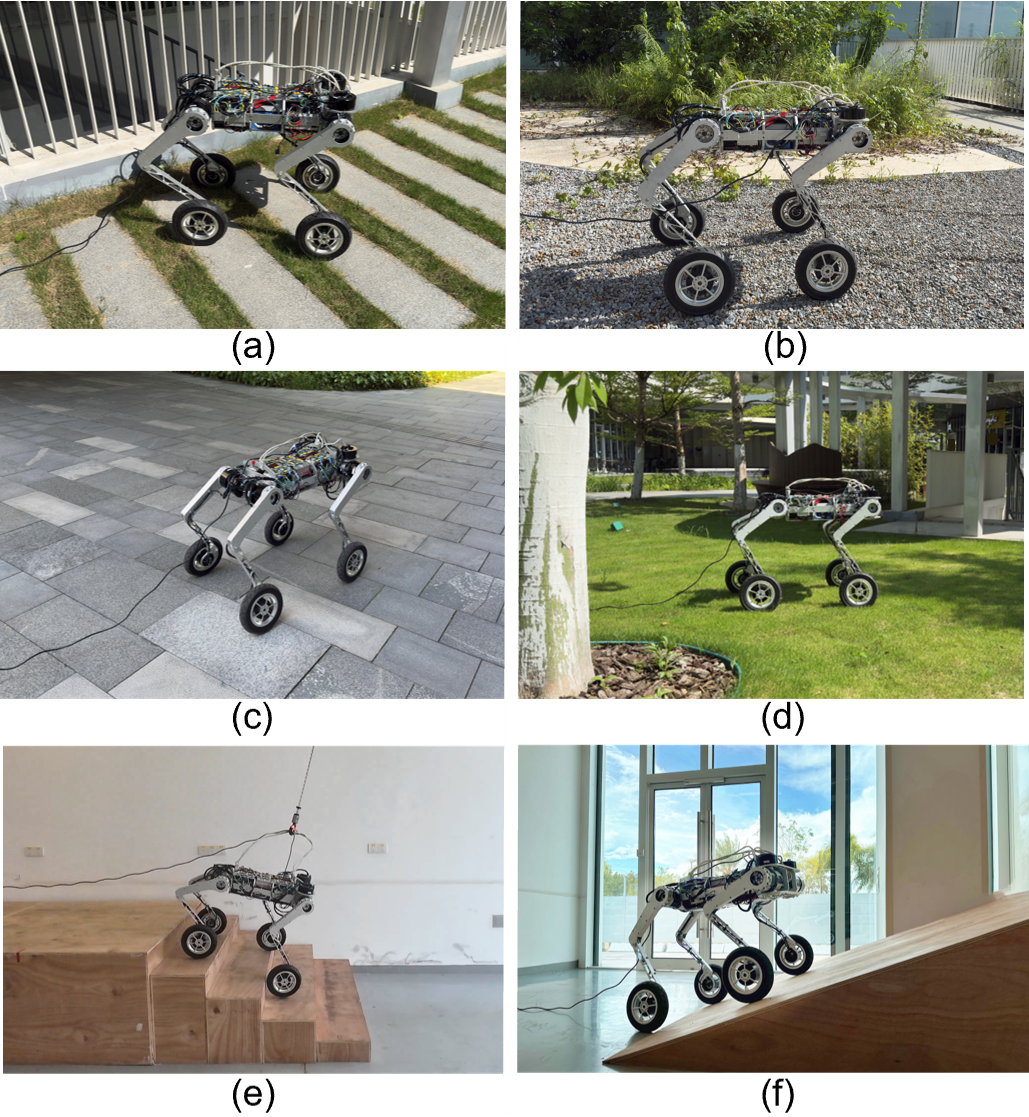}
    \caption{By utilizing the modified hip-yaw DOF, \flores demonstrates efficient and versatile locomotion across various terrains, including (a) discrete terrain, (b) gravel, (c) paved surfaces, (d) grassland, (e) stairs, and (f) slopes. Supplementary videos are available on our project website.}
    \label{fig:terrain}
    \vspace{-0.5cm}
\end{figure}

\flores demonstrates efficient real-world locomotion across diverse terrains, as illustrated in Fig.~\ref{fig:staris} and Fig.~\ref{fig:terrain}. 
The evaluation encompasses unstructured terrains, including discrete ground (Fig. \ref{fig:terrain}(a)), gravel (Fig. \ref{fig:terrain}(b)), paved surfaces (Fig. \ref{fig:terrain}(c)), grassland (Fig. \ref{fig:terrain}(d)), and slopes (Fig. \ref{fig:terrain}(f)), as well as stairs with a maximum height of 15 cm (Fig. \ref{fig:terrain}(e)). 
Fig. \ref{fig:staris} shows the robot's joint angle paths for the front thigh and calf joints, illustrating its response when encountering stairs.

As shown in Fig. \ref{fig:staris}, \flores begins in driving mode on a flat surface. 
When a stumble caused by the stairs is detected through proprioceptive feedback, \flores seamlessly transits into walking mode to ascend the steps. 
Upon completing the ascent, the robot automatically reverts to driving mode, enhancing its navigational efficiency.

While traversing different terrains, as illustrated in Fig.~\ref{fig:terrain}, the customized learning policy allows \flores to automatically adopt different locomotion strategies that effectively leverage its unique mechanical structure, resulting in lower energy consumption and facilitating high-efficiency movement across various terrains and tasks, which we will further validate in the following section.

\subsection{Comparison of {CoT}}

Our experiments validate the enhanced steering ability and navigation efficiency of \flores across various scenarios, including straight-line motion, turning maneuvers, and complex tasks requiring continuous directional adjustments.
These scenarios reflect common operating conditions for mobile robots in unstructured terrains.

Additionally, the data collected from B2W serves as the benchmark. 
By contrasting the performance metrics of \flores with those of B2W, we aim to highlight the advantages of our robot in terms of energy efficiency and maneuverability. 
This comparison underscores the effectiveness of \flores in navigating diverse environments, further demonstrating its capabilities in real-world applications.

To quantify the efficiency, we introduce CoT, a key metric for assessing energy efficiency in robotic movement.
Following \cite{doi:10.1126/scirobotics.adi9641}, the mechanical CoT is defined as
\begin{equation}
    \textup{CoT} = \sum_{\textup{all joints}}[\tau\dot{\theta}]^{+}/(mg|v^{b}_{xy}|),
    \label{eq:cot}
\end{equation}
where $\tau$ and $\dot{\theta}$ are the torque and speed of the joint, $[\cdot]^+$ is the positive term of the product, $mg$ is the weight of the robot, and $|v^{b}_{xy}|$ is the horizontal speed of the robot's base. In indoor experiments, \( |v^{b}_{xy}| \) is obtained from the motion capture system provided by OptiTrack, while in outdoor environments, it is calculated by averaging the speed during traversal.
The B2W wheel modules exhibit an additional Coulomb-like friction torque offset $\tau_{\mathrm{seal}}$ induced by the waterproof sealing structure.
To avoid biasing the wheel-side power, we subtract this sealing-induced offset before computing CoT.
Let $\mathcal{W}$ denote the set of wheel actuators, and the corrected CoT is defined as
\begin{equation}
    \mathrm{CoT}_{\text{corr}} =
    \frac{
      \sum_{j \notin \mathcal{W}}[\tau_j \dot{\theta}_j]^+
      \;+\;
      \sum_{j \in \mathcal{W}}\bigl[\tau_j \dot{\theta}_j - \tau_{\mathrm{seal}}\lvert \dot{\theta}_j\rvert \bigr]^+
    }{m g\,\lvert v^{b}_{xy}\rvert}.
\end{equation}
Note that $\tau_{\mathrm{seal}}$ is measured quasi-statically with a calibrated torque wrench.
Unless otherwise specified, all B2W CoT values reported below are computed with the above wheel-sealing friction compensation applied to the wheel joints, while \flores and all non-wheel joints use the uncorrected term in \eqref{eq:cot}.

\subsubsection{Straight-Line Motion}
\begin{figure}[tbp]
    \centering
    \includegraphics[width = 0.48\textwidth]{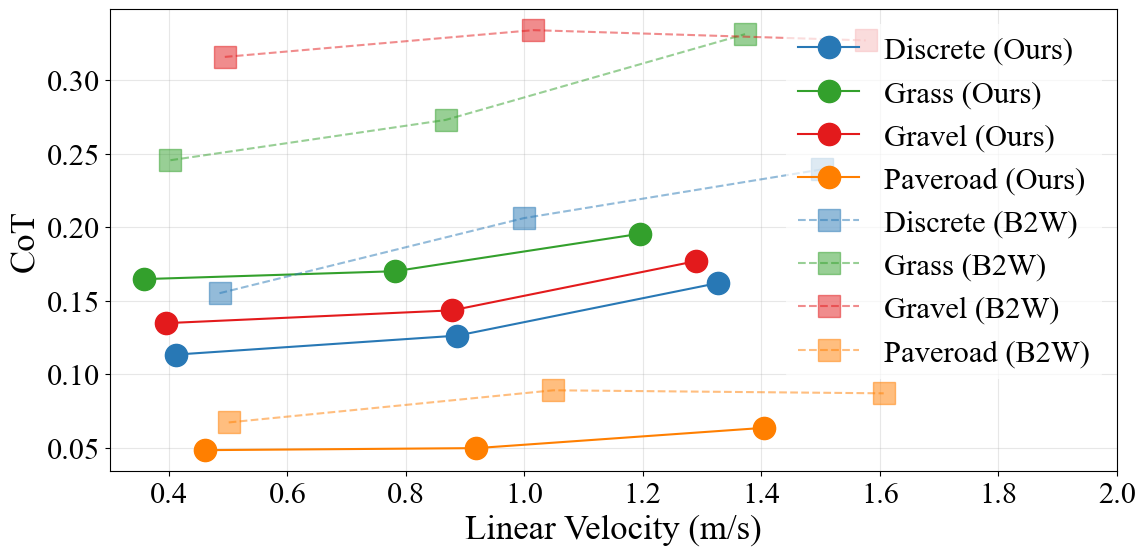}
    \caption{Comparison of CoT for \flores and B2W during straight-line motion at different velocities across various terrains.}
    \label{fig:acc_cot}
    \vspace{-0.5cm}
\end{figure}

We conduct straight-line motion experiments on the paved ground, grassland, discrete ground, and gravel terrain with \flores and B2W.
The robots are evaluated at speeds ranging from [0.5\,m/s, 1.5\,m/s]. 
The joint torque $\tau$ and joint velocity $\dot{\theta}$ are directly obtained from the joint encoder.
During planar straight-line tracking, where the robot is required to maintain a consistent heading while rolling forward, the front hip-yaw joint stays almost stationary and applies small corrective torques as needed; the magnitude and frequency of these pulses increase with terrain disturbance.
In contrast, the typical hip-roll joint needs to continuously provide torque to maintain the leg at the proper position.
The resulting CoT is shown in Fig.~\ref{fig:acc_cot}, where \flores achieves the lowest CoT on all unstructured terrains.

\subsubsection{Lateral Walking}
\begin{figure}[tbp]
    \centering
    \includegraphics[width = 0.48\textwidth]{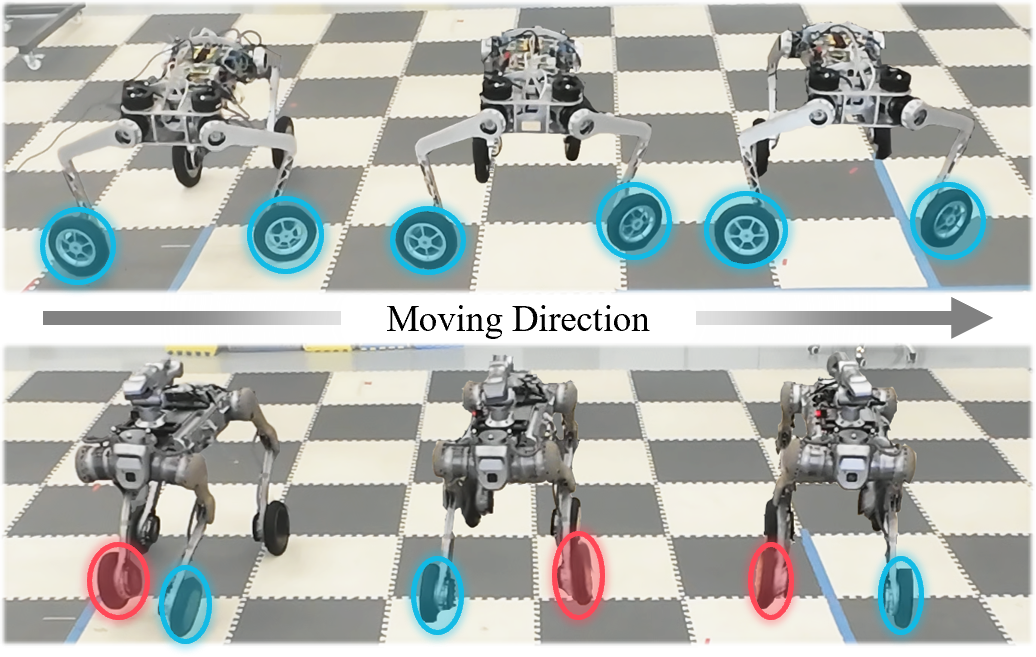}
    \caption{\flores and B2W are performing lateral walking. The blue circles indicate the wheels in contact with the ground, while the red circles represent the wheels that are lifted off the ground. The modified hip-yaw degree of freedom in \flores allows it to utilize the front wheels to assist in lateral movement. In contrast, B2W must continuously lift its legs to achieve lateral motion, resulting in a less efficient locomotion strategy.} 
    \label{fig:lateral}
    \vspace{-0.6cm}
\end{figure}

Comparisons of locomotion and CoT between \flores and B2W for lateral walking are illustrated in Fig.~\ref{fig:lateral}. 
The modification enables the leg structure to rotate so that the front wheel can directly move in the lateral direction, while the rear legs continue to function in a typical lifting motion to facilitate movement. 
In contrast, the B2W robot requires all legs to be lifted and lowered continuously during lateral movement, which negatively affects the efficiency of its wheeled structure. 
We test lateral movement for both \flores and B2W at a commanded speed of \(0.5\,\text{m/s}\). \flores achieves a CoT of 0.3614, which is significantly superior to B2W's CoT of 0.6767.

\subsubsection{Turning Maneuvers}
\begin{figure}[tbp]
    \centering
    \includegraphics[width = 0.45\textwidth]{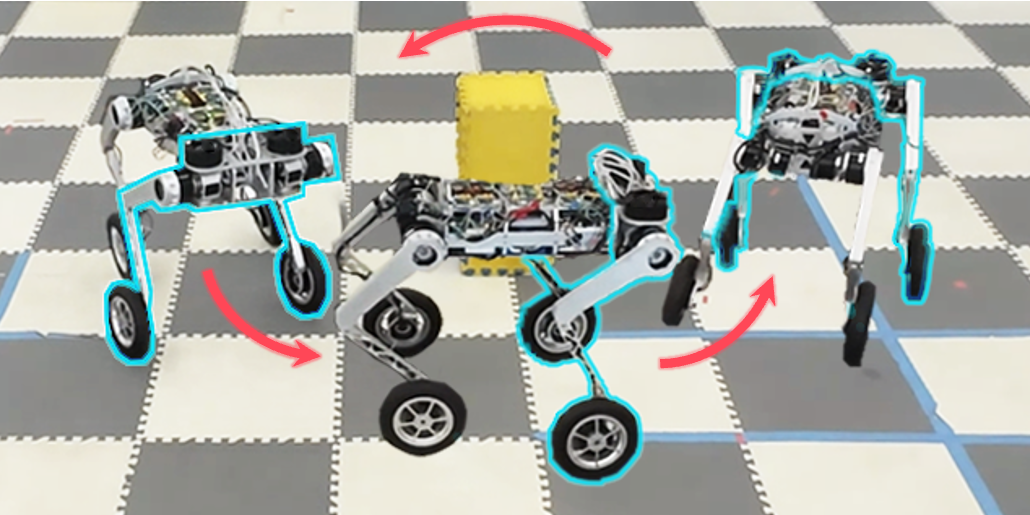}
    \caption{\flores is commanded to follow circular paths with a radius of 0.5 m, 1.0 m, 1.5 m, and 2.0 m. The modified hip-yaw DoF, highlighted by the blue line, serves as a steering mechanism that enables \flores to change its heading angle while keeping the front legs on the ground. This allows the efficient turning without lifting the front legs, thereby reducing overall energy consumption.}
    \label{fig:turning}
    \vspace{-0.3cm}
\end{figure}

To validate the effectiveness of \flores's steering capabilities, we compare the CoT between \flores and B2W while following circular paths of varying radius. 
As shown in Fig.~\ref{fig:turning}, the robots follow circular paths with a radius ranging from [0.5\,m, 2\,m], and travel at a constant input forward velocity $v_x= 0.4\,\textup{m/s}$.
Using measured yaw from OptiTrack, the PID controller maps the tracking error to the desired angular velocity $\omega_{z}^{\text{cmd}}$ for the RL policy.
During these path-following tasks, \flores utilizes the modified hip-yaw DoF as its steering mechanism to adjust its heading direction. 
In this maneuver, only the rear legs of \flores need to be lifted while the front legs remain in contact with the ground. 
In contrast, B2W needs to lift all its legs to continuously change its heading direction, resulting in significantly higher energy expenditure.
Table \ref{tab:CoT_of_turning} illustrates that the smaller the turning radius, the more pronounced the advantages of \flores become in terms of efficiency.
This is because a smaller radius necessitates more rapid changes in the robot's heading direction. 
\flores can accommodate this requirement effectively by simply using a larger steering angle provided by its modified hip-yaw DoF. 
However, B2W must lift and lower its legs at a higher frequency, which leads to increased energy consumption and reduced efficiency during turns.

\subsubsection{Path Following}
\begin{figure}[t]
    \centering
    \includegraphics[width = 0.48\textwidth]{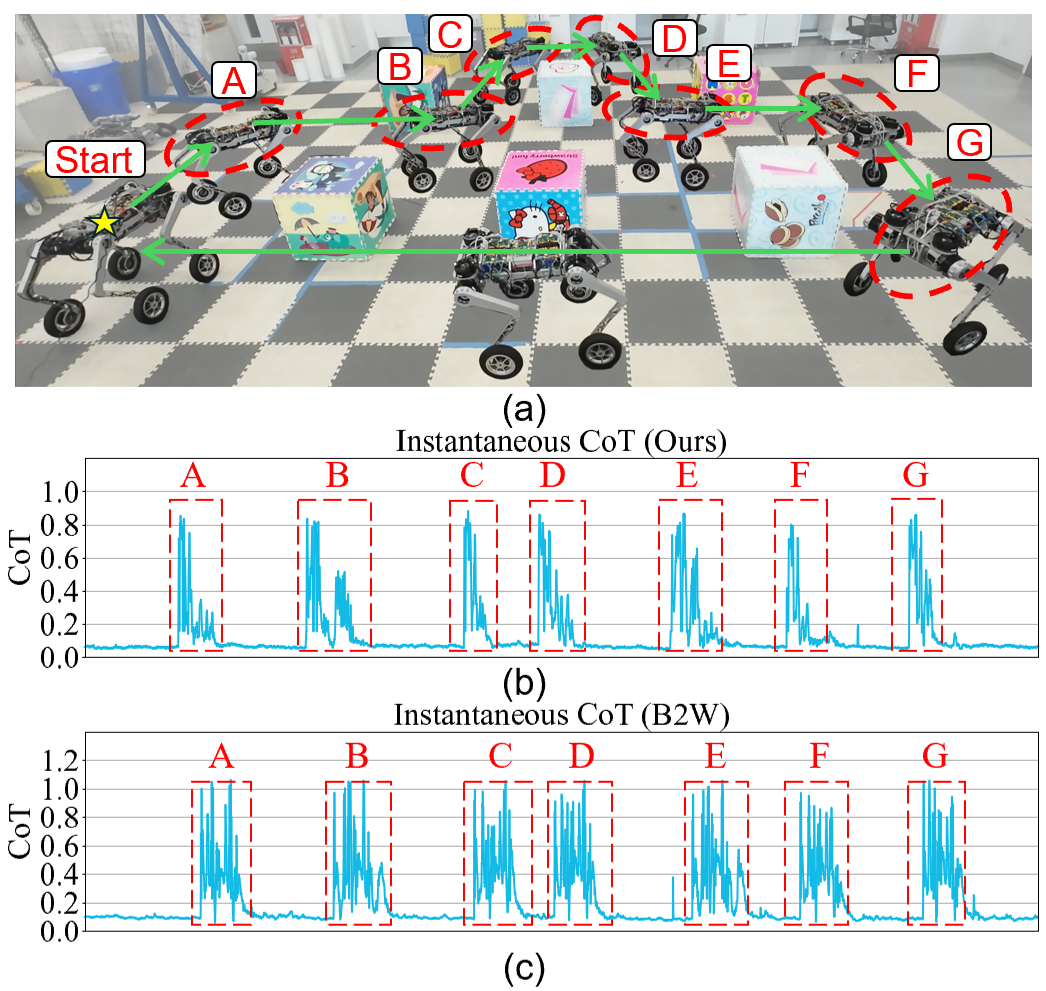}
    \caption{\flores is executing a complex path-following task that necessitates continuous and rapid changes in direction. (a) illustrates the path, with points A, B, C, D, E, F, and G marking the seven points where the robot rapidly changes its direction. (b) and (c) display the instantaneous CoT for \flores and B2W as they navigate the same path-following task. The spikes in the plots correspond to the turning points, clearly demonstrating that \flores achieves greater effectiveness in this task.}
    \label{fig:path_and_cot}
    \vspace{-0.5cm}
\end{figure}

After validating straight-line and turning motions, we test a complex path combining both motions to evaluate the navigation efficiency of \flores. 
The robot is required with maneuvering through six cluttered boxes, as depicted in Fig. \ref{fig:path_and_cot}. 
At points A, B, C, D, E, F, and G, the robot is required to make 90 degrees turns.
Fig. \ref{fig:path_and_cot}(b) and (c) illustrate the instantaneous CoT for both \flores and B2W while following the path. 
Notably, at each turning point, \flores's CoT was approximately 70\% of that of B2W, and this highlights the superior efficiency of \flores in navigating through the cluttered environment.

\begin{table}[tbp]
  \centering
  \caption{Comparison of CoT for turning maneuvers}
    \begin{tabular}{cccc}
    \toprule
    \multirow{2}[2]{*}{Radius} & \multicolumn{1}{c}{\multirow{2}[2]{*}{CoT (\flores)}} & \multicolumn{1}{c}{\multirow{2}[2]{*}{CoT (B2W)}} \\
          &       &         \\
    \midrule
    0.5\,m & \textbf{0.24}  & 0.357  \\
    \midrule
    1.0\,m & \textbf{0.18}  & 0.256 \\
    \midrule
    1.5\,m & \textbf{0.149} & 0.195  \\
    \midrule
    2.0\,m & \textbf{0.139} & 0.183    \\
    \bottomrule
    \end{tabular}%
  \label{tab:CoT_of_turning}%
  \vspace{-10pt}
\end{table}%


\section{CONCLUSIONS}
This work presents a novel wheel-legged robot, \flores, featuring an innovative front-leg design that synthesizes the speed and efficiency of wheels with the versatility and adaptability of legs, which significantly enhances the performance on unstructured terrain.
\flores offers enhanced steering ability, and it is well-suited for dynamic hybrid locomotion in diverse scenarios with its high output-speed joint modules.
Additionally, we propose an RL-based control framework, with a customized reward mechanism tailored to the unique mechanical design of \flores.
Through real-world experiments, \flores demonstrates noteworthy flexibility and efficiency across various environments, validating the effectiveness of the proposed design and control strategy.





\bibliographystyle{IEEEtran}
\bibliography{ref}

\end{document}